\RequirePackage{fix-cm}
\documentclass[twocolumn]{svjour3}          
\smartqed  
\usepackage{graphicx}
\usepackage{url}
\usepackage[colorlinks = true]{hyperref}
\usepackage{amsmath,amssymb,amsbsy}
\usepackage{paralist}
\usepackage{xcolor}
\usepackage{color}
\usepackage{graphicx}
\graphicspath{{./figs/}}
\usepackage{algorithm}
\usepackage{algorithmic}
\usepackage{comment}
\usepackage{fancyhdr}
\usepackage{cite}
\usepackage{cleveref}
\usepackage{enumitem}

\newtheorem{thm}{Theorem}
\newtheorem{cor}{Corollary}
\newtheorem{defi}{Definition}

\newtheorem{lem}{Lemma}

\newcommand{\R}{\mathbb{R}}

\newcommand{\N}{\mathbb{N}}

\newcommand{\e}{\begin{equation}}
\newcommand{\ee}{\end{equation}}
\newcommand{\en}{\begin{equation*}}
\newcommand{\een}{\end{equation*}}
\newcommand{\eqn}{\begin{eqnarray}}
\newcommand{\eeqn}{\end{eqnarray}}
\newcommand{\bmat}{\begin{bmatrix}}
\newcommand{\emat}{\end{bmatrix}}

\newcommand{\vct}[1]{\boldsymbol{#1}}
\newcommand{\mtx}[1]{\boldsymbol{#1}}

\newcommand{\T}{\mathrm{T}}

\newcommand{\trace}{\operatorname{trace}}

\newcommand{\rank}{\operatorname{rank}}

\newcommand{\set}[1]{\mathcal{#1}}

\DeclareMathOperator*{\argmin}{\text{arg~min}}

\newcommand{\calC}{\mathcal{C}}

\newcommand{\calX}{\mathcal{X}}

\newcommand{\va}{\vct{a}}
\newcommand{\vb}{\vct{b}}

\newcommand{\vp}{\vct{p}}
\newcommand{\vq}{\vct{q}}

\newcommand{\vx}{\vct{x}}

\newcommand{\vz}{\vct{z}}
\newcommand{\valpha}{\vct{\alpha}}

\newcommand{\vphi}{\vct{\phi}}
\newcommand{\vpsi}{\vct{\psi}}

\newcommand{\vzero}{\vct{0}}

\newcommand{\mA}{\mtx{A}}
\newcommand{\mB}{\mtx{B}}

\newcommand{\mL}{\mtx{L}}

\newcommand{\mP}{\mtx{P}}
\newcommand{\mQ}{\mtx{Q}}
\newcommand{\mR}{\mtx{R}}

\newcommand{\mU}{\mtx{U}}
\newcommand{\mV}{\mtx{V}}
\newcommand{\mW}{\mtx{W}}
\newcommand{\mX}{\mtx{X}}
\newcommand{\mY}{\mtx{Y}}
\newcommand{\mZ}{\mtx{Z}}

\newcommand{\mDelta}{\mtx{\Delta}}
\newcommand{\mLambda}{\mtx{\Lambda}}
\newcommand{\mOmega}{\mtx{\Omega}}
\newcommand{\mPhi}{\mtx{\Phi}}
\newcommand{\mPsi}{\mtx{\Psi}}
\newcommand{\mPi}{\mtx{\Pi}}
\newcommand{\mSigma}{\mtx{\Sigma}}
\newcommand{\mTheta}{\mtx{\Theta}}

\newcommand{\mId}{{\bf I}}

\newcommand{\mzero}{{\bf 0}}

\newcommand{\setC}{\set{C}}

\newcommand{\setO}{\set{O}}

\newcommand{\setX}{\set{X}}

\graphicspath{{./figs/}}

\newlength{\imgwidth}
\newboolean{twoColVersion}
\setboolean{twoColVersion}{false}
\newcommand{\twoCol}[2]{\ifthenelse{\boolean{twoColVersion}} {#1} {#2} }

\begin{document}

\title{The Global Optimization Geometry of Shallow Linear Neural Networks\thanks{ZZ and MBW were supported by NSF grant CCF--1409261, NSF CAREER grant CCF--1149225, and the DARPA Lagrange Program under ONR/SPAWAR contract N660011824020. DS was supported by the Israel Science Foundation (grant No. 31/1031), and by the Taub Foundation. }}


\author{Zhihui Zhu        \and
       Daniel Soudry \and Yonina C. Eldar \and Michael B. Wakin
}


\institute{Z. Zhu \at
	Mathematical Institute for Data Science\\Johns Hopkins University\\
             \email{zzhu29@jhu.edu}
           \and
         D. Soudry and Y. C. Eldar \at
            Department of Electrical Engineering\\Technion, Israel Institute of Technology\\
            \email{daniel.soudry@technion.ac.il}\\
            \email{yonina@ee.technion.ac.il}
            \and
          M. B. Wakin  \at
Department of Electrical Engineering\\Colorado School of Mines\\
\email{mwakin@mines.edu.}
}

\date{Received: date / Accepted: date}

\maketitle

\begin{abstract}
{ We examine the squared error loss landscape of shallow linear neural networks. 
We show---with  significantly milder assumptions than previous works---that the corresponding optimization problems have benign geometric properties: there are no spurious local minima and the Hessian at every saddle point has at least one negative eigenvalue. This means that at every saddle point there is a directional negative curvature which algorithms can utilize to further decrease the objective value.  These geometric properties imply that many local search algorithms (such as the gradient descent which is widely utilized for training neural networks) can provably solve the training problem with global convergence.
}

\end{abstract}

\section{Introduction}
A neural network consists of a sequence of operations (a.k.a. layers), each of which performs a linear transformation of its input, followed by a point-wise activation function, such as a sigmoid function or the rectified linear unit (ReLU)~\cite{schalkoff1997artificial}. Deep artificial neural networks (i.e., deep learning) have recently led to state-of-the-art empirical performance in many areas including computer vision, machine learning, and signal processing~\cite{hornik1989multilayer,lecun2015deep,mousavi2015deep,kamilov2016learning,li2016unsupervised,borgerding2017amp,li2018learning}.

One crucial property of neural networks is their ability to approximate nonlinear functions. It has been shown that even a shallow neural network (i.e., a network with only one hidden layer) with a point-wise activation function has the universal approximation ability~\cite{hornik1989multilayer,cybenko1989approximation}. In particular, a shallow network with a sufficient number of activations (a.k.a. neurons) can approximate continuous functions on compact subsets of $\R^{d_0}$ with any desired accuracy, where $d_0$ is the dimension of the input data.

However, the universal approximation theory does not guarantee the algorithmic learnability of those parameters which correspond to the linear transformation of the layers. Neural networks may be trained (or learned) in an unsupervised manner, a semi-supervised manner, or a supervised manner which is by far the most common scenario. With supervised learning, the neural networks are trained by minimizing a loss function in terms of the parameters to be optimized and the training examples that consist of both input objects and the corresponding outputs. A popular approach for optimizing or tuning the parameters is gradient descent with the backpropagation method efficiently computing the gradient~\cite{werbos1974beyond}.

Although gradient descent and its variants work surprisingly well for training neural networks in practice, it remains an active research area to fully understand the theoretical underpinnings of this phenomenon. In general, the training optimization problems are nonconvex and it has been shown that even training a simple neural network is NP-complete {in general}~\cite{blum1989training}. There is a large and rapidly increasing literature on the optimization theory of neural networks, surveying all of which is well outside our scope. Thus, we only
briefly survey the works most relevant to ours.

In seeking to better understand the optimization problems in training neural networks, one line of research attempts to analyze their {\em geometric landscape}. The geometric landscape of an objective function relates to questions concerning the existence of spurious local minima and the existence of negative eigenvalues of the Hessian at saddle points. If the corresponding problem has no spurious local minima and all the saddle points are {\em strict} (i.e., the Hessian at any saddle point has a negative eigenvalue), then a number of local search algorithms~\cite{lee2016gradient,ge2015escaping,jin2017escape,lee2017first} are guaranteed to find globally minimal solutions. Baldi and Hornik~\cite{baldi1989neural} showed that there are no spurious local minima in training shallow linear neural networks but did not address the geometric landscape around saddle points. Kawaguchi~\cite{kawaguchi2016deep} further extended the analysis in~\cite{baldi1989neural} and showed that the loss function for training a general linear neural network has no spurious local minima and satisfies the strict saddle property (see Definition~\ref{def:strict saddle property} in Section~\ref{sec:Preliminaries}) for shallow neural networks under certain conditions. Kawaguchi also proved that for general deeper networks, there exist saddle points at which the Hessian is positive semi-definite (PSD), i.e., does not have any negative eigenvalues.

With respect to nonlinear neural networks, it was shown that there are no spurious local minima for a network with one ReLU node~\cite{tian2017analytical,du2017convolutional}. However, it has also been proved that there do exist spurious local minima in the population loss of shallow neural networks with even a small number (greater than one) of ReLU activation functions~\cite{safran2017spurious}. Fortunately, the number of spurious local minima can be significantly reduced with an over-parameterization scheme~\cite{safran2017spurious}. Soudry and Hoffer~\cite{soudry2017exponentially} proved that the number of sub-optimal local minima is negligible compared to the volume of global minima for multilayer neural networks when the number of training samples $N$ goes to infinity and the number of parameters is close to $N$. Haeffele and Vidal~\cite{haeffele2015global} provided sufficient conditions to guarantee that certain local minima (having an all-zero slice) are also global minima. The training loss of multilayer neural networks at differentiable local minima was examined in~\cite{soudry2016no}. Yun~et al.~\cite{yun2017global} very recently provided sufficient and necessary conditions to guarantee that certain critical points are also global minima.

\setlength{\tabcolsep}{3pt}
\begin{table*}[htp!]\caption{Comparison of different results on characterizing the geometric lanscape of the objective function in training a shallow linear network (see \eqref{eq:shallow linear network}). Here, ? means this point is not discussed {and \checkmark\textsuperscript{\kern-0.55em x} indicates the result covers degenerate critical points only.} }\label{table:summary}
	\begin{center}
		\small
		\renewcommand{\arraystretch}{1.4}
		\begin{tabular}{c|c|c|c|c}
			\hline result & regularizer & condition & \begin{tabular}{c} no spurious\\ local minima\end{tabular} & \begin{tabular}{c}strict saddle\\ property\end{tabular} \\
			\hline \hline \cite[Fact 4]{baldi1989neural} & no & \begin{tabular}{c}$\mX\mX^\T$ and $\mY\mY^\T$ are of full row rank, $d_2 \leq d_0$, \\ $\mY\mX^\T (\mX\mX^\T)^{-1}\mX\mY^\T$ has $d_2$ distinct eigenvalues\end{tabular} & \checkmark & ? \\
			\hline
			\cite[Theorem 2.3]{kawaguchi2016deep} & no & \begin{tabular}{c}$\mX\mX^\T$ and $\mY\mY^\T$ are of full row rank, $d_2 \leq d_0$, \\ $\mY\mX^\T (\mX\mX^\T)^{-1}\mX\mY^\T$ has $d_2$ distinct eigenvalues\end{tabular} & \checkmark & \checkmark\\
			\hline \cite[Theorem 2.1]{lu2017depth}& no & $\mX\mX^\T$ and $\mY\mY^\T$ are of full row rank & \checkmark & ? \\
			\hline \cite[Theorem 1]{laurent2017deep} & no & $d_1 \geq \min(d_0,d_2)$ & \checkmark & ?  \\
					\hline 	\cite[Theorem 8]{nouiehed2018learning} & no & no & \checkmark &  \checkmark\textsuperscript{\kern-0.55em x}  \\	
			\hline \cite[Theorem 3]{zhu2017GlobalOptimality} & $\|\mW_2^\T \mW_2 - \mW_1 \mW_1^\T\|_F^2$ &  $d_1 \leq \min(d_0, d_2)$ and conditions \eqref{eq:cond low-rank for network} and \eqref{eq:rank constraint} & \checkmark & \checkmark\\
			\hline \Cref{thm:low-rank for network 2} & $\|\mW_2^\T \mW_2 - \mW_1 \mX \mX^\T \mW_1^\T\|_F^2$ & $\mX\mX^\T$ is of full row rank &  \checkmark & \checkmark\\
			\hline \Cref{thm:low-rank for network no reg} &  no &  $\mX\mX^\T$ is of full row rank &  \checkmark & \checkmark\\
			\hline
		\end{tabular}
	\end{center}
\end{table*}

A second line of research attempts to understand the reason that local search {{\em algorithms}} efficiently find a local minimum. Aside from standard Newton-like methods such as cubic regularization~\cite{nesterov2006cubic} and the trust region algorithm~\cite{byrd2000trust}, recent  work~\cite{lee2016gradient,ge2015escaping,jin2017escape,lee2017first} has shown that first-order methods also efficiently avoid strict saddles. It has been shown in~\cite{lee2016gradient,lee2017first} that a set of first-order local search techniques (such as gradient descent) with random initialization almost surely avoid strict saddles. Noisy gradient descent~\cite{ge2015escaping} and a variant called perturbed gradient descent~\cite{jin2017escape} have been proven to efficiently avoid strict saddles from any initialization. Other types of algorithms utilizing second-order (Hessian) information~\cite{agarwal2016finding,carmon2016accelerated,curtis2017exploiting} can also efficiently find approximate local minima.

To guarantee that gradient descent type algorithms (which are widely adopted in training neural networks) converge to the global solution, the behavior of the saddle points of the objective functions in training neural networks is as important as the behavior of local minima.\footnote{ {From an optimization perspective, non-strict saddle points and local minima have similar first-/second-order information and it is hard for first-/second-order methods (like gradient descent) to distinguish between them.}}  However, the former has rarely been investigated compared to the latter, even for shallow linear networks. It has been shown in~\cite{baldi1989neural,kawaguchi2016deep,lu2017depth,laurent2017deep} that the objective function in training shallow linear networks has no spurious local minima under certain conditions. The behavior of saddle points is considered in~\cite{kawaguchi2016deep}, where the strict saddle property is proved  for the case where both the input objects $\mX\in \R^{d_0\times N}$ and the corresponding outputs $\mY\in\R^{d_2\times N}$ of the training samples have full row rank, $\mY\mX^\T (\mX\mX^\T)^{-1}\mX\mY^\T$ has distinct eigenvalues, and $d_2 \leq d_0$. While the assumption on $\mX$ can be easily satisfied, the assumption involving $\mY$ implicity adds constraints on the true weights. Consider a simple case where $\mY = \mW_2^\star\mW_1^\star\mX$, with $\mW_2^\star$ and $\mW_1^\star$  the underlying weights to be learned. Then the full-rank assumption on $\mY\mY^{\T} = \mW_2^\star\mW_1^\star\mX\mX^\T \mW_1^{\star\T}\mW_2^{\star\T}$ at least requires $\min(d_0,d_1)\geq d_2$ and $\rank(\mW_2^\star\mW_1^\star)\geq d_2$.  {Recently, the strict saddle property was also shown to hold without the above conditions on $\mX$, $\mY$, $d_0$, and $d_2$, but only for {\em degenerate} critical points, specifically those points where $\rank(\mW_2\mW_1)<\min\{d_2,d_1,d_0\}$~\cite[Theorem 8]{nouiehed2018learning}.}

In this paper we analyze the optimization geometry of the loss function in training shallow linear neural networks. In doing so, we first characterize the behavior of {{\em all}} critical points of the corresponding optimization problems with an additional regularizer (see \eqref{low-rank for network 2}), but without requiring the conditions used in~\cite{kawaguchi2016deep}  except the one on the input data $\mX$.  In particular, we examine
the loss function for training a shallow linear neural network with an additional regularizer and show that it has no spurious local minima and obeys the strict saddle property if the input $\mX$ has full row rank.  This benign geometry ensures that a number of local search algorithms---including gradient descent---converge to a global minimum when training a shallow linear neural network with the proposed regularizer.  We note that the additional regularizer (in~\eqref{low-rank for network 2}) is utilized to shrink the set of critical points and has no effect on the global minimum of the original problem.  We also observe from experiments that this additional regularizer speeds up the convergence of iterative algorithms in certain cases.
{Building on our study of the regularized problem and on~\cite[Theorem 8]{nouiehed2018learning}, we then show that these benign geometric properties are preserved even {\em without} the additional regularizer under the same assumption on the input data.
}
Table~\ref{table:summary} summarizes our main result and those of { related works} on characterizing the geometric landscape of the loss function in training shallow linear neural networks. 

Outside of the context of neural networks, such geometric analysis (characterizing the behavior of all critical points) has been recognized as a powerful tool for understanding nonconvex optimization problems in applications such as phase retrieval \cite{sun2016geometric,qu2017convolutional}, dictionary learning \cite{sun2015complete}, tensor factorization~\cite{ge2015escaping}, phase synchronization~\cite{liu2016estimation} and
low-rank matrix optimization~\cite{bhojanapalli2016lowrankrecoveryl,park2016non,li2017geometry,ge2016matrix,li2016symmetry,zhu2017global,zhu2017GlobalOptimality,ge2017no}. A similar regularizer (see \eqref{low-rank for network}) to the one used in~\eqref{low-rank for network 2} is also utilized in~\cite{park2016non,li2017geometry,zhu2017global,zhu2017GlobalOptimality,ge2017no} for analyzing the optimization geometry.

The outline of this paper is as follows. Section~\ref{sec:Preliminaries} contains the formal definitions for strict saddles and the strict saddle property. Section~\ref{sec:main results} presents our main result on the geometric properties for training shallow linear neural networks.  The proof of our main result is given in \Cref{sec:prf}.

\section{Preliminaries}\label{sec:Preliminaries}
\subsection{Notation}

We use the symbols $\mId$ and $\mzero$ to respectively represent the identity matrix and zero matrix with appropriate sizes. We denote the set of $r\times r$ orthonormal matrices  by $\setO_r:=\{\mR\in\R^{r\times r}:\mR^\T\mR = \mId\}$. If a function $g(\mW_1,\mW_2)$ has two arguments, $\mW_1\in\R^{d_1\times d_0}$ and $\mW_2\in\R^{d_2\times d_1}$, then we occasionally use the notation $g(\mZ)$ where we stack these two matrices into a larger one via $\mZ=\begin{bmatrix}\mW_2 \\ \mW_1^\T \end{bmatrix}$. For a scalar function $h(\mW)$ with a matrix variable $\mW\in\R^{d_2\times d_0}$, its gradient is a $d_2\times d_0$ matrix whose $(i,j)$-th entry is $[\nabla h(\mW)]_{ij} = \frac{\partial h(\mW)}{\partial W_{ij}}$ for all $i\in [d_2], j\in [d_0] $. Here $[d_2] = \{1,2,\ldots,d_2\}$ for any $d_2\in \N$ and $W_{ij}$ is the $(i,j)$-th entry of the matrix $\mW$.
Throughout the paper, the Hessian of $h(\mW)$ is represented by a bilinear form defined via $[\nabla^2 h(\mW)](\mA,\mB) = \sum_{i,j,k,l}\frac{\partial^2 h(\mW)}{\partial W_{ij}\partial W_{kl}} A_{ij} B_{kl}$ for any $\mA,\mB\in\R^{d_2\times d_0}$. Finally, we use $\lambda_{\min}(\cdot)$ to denote the smallest eigenvalue of a matrix.

\subsection{Strict saddle property}
Suppose $h:\R^{n}\rightarrow \R$ is a twice continuously differentiable objective function. The notions of critical points, strict saddles, and the strict saddle property are formally defined as follows.

\begin{defi}[Critical points]
	$\vx$ is called a critical point of $h$ if the gradient at $\vx$ vanishes, i.e., $\nabla h(\vx) = \mzero$.
\end{defi}

\begin{defi}[Strict saddles~\cite{ge2015escaping}]
	We say a critical point $\vx$ is a strict saddle if the Hessian evaluated at this point has at least one strictly negative eigenvalue, i.e., $\lambda_{\min}(\nabla^2 h(\vx))<0$.
\end{defi}

In words, for a strict saddle, also called a ridable saddle~\cite{sun2015complete}, its Hessian has at least one negative eigenvalue which implies that there is a directional negative curvature that algorithms can utilize to further decrease the objective value. This property ensures that many local search algorithms can escape strict saddles by either directly exploiting the negative curvature~\cite{curtis2017exploiting} or adding noise which serves { as} a surrogate of the negative curvature~\cite{ge2015escaping,jin2017escape}. On the other hand, when a saddle point has a Hessian that is positive semidefinite (PSD), it is difficult for first- and second-order methods to avoid converging to such a point. In other words, local search algorithms require exploiting higher-order (at least third-order) information in order to escape from a critical point that is neither a local minimum nor a strict saddle. We note that any local maxima are, by definition, strict saddles.

The following strict saddle property defines a set of nonconvex functions that can be efficiently minimized by a number of iterative algorithms with guaranteed convergence.
\begin{defi}[Strict saddle property~\cite{ge2015escaping}]\label{def:strict saddle property}
	A twice differentiable function satisfies the strict saddle property if each critical point either corresponds to a local minimum or is a strict saddle.
\end{defi}

Intuitively, the strict saddle property requires a function to have a negative curvature direction---which can be exploited by a number of iterative algorithms such as noisy gradient descent~\cite{ge2015escaping} and the trust region method~\cite{conn2000trust} to further decrease the function value---at all critical points except for local minima.

\begin{thm}\label{thm:global convergence}
	\cite{nesterov2006cubic,sun2015complete,ge2015escaping,lee2016gradient}
	For a twice continuously differentiable objective function satisfying the strict saddle property, a number of iterative optimization algorithms  can find a local minimum. In particular, for such functions,
	\begin{itemize}[label=\tiny\textbullet]
		\item gradient descent almost surely converges to a local minimum with a random initialization~\cite{ge2015escaping};
		\item noisy gradient descent~\cite{ge2015escaping} finds a local minimum with high probability and any initialization; and
		\item Newton-like methods such as cubic regularization~\cite{nesterov2006cubic} converge to a local minimum with any initialization.
	\end{itemize}
\end{thm}

\Cref{thm:global convergence} ensures that many local search algorithms can be utilized to find a local minimum for strict saddle functions (i.e., ones obeying the strict saddle property). This is the main reason that significant effort has been devoted to establishing the strict saddle property for different problems~\cite{kawaguchi2016deep,sun2016geometric,qu2017convolutional,park2016non,li2017geometry,zhu2017global}.

In our analysis, we further characterize local minima as follows.
\begin{defi}[Spurious local minima]\label{def:strict saddle property}
	We say a critical point $\vx$ is a spurious local minimum if it is a local minimum but not a global minimum.
\end{defi}
In other words, we separate the set of local minima into two categories: the global minima and the spurious local minima which are not global minima. Note that most local search algorithms are only guaranteed to find a local minimum, which is not necessarily a global one. Thus, to ensure the local search algorithms listed in \Cref{thm:global convergence} find a global minimum, in addition to the strict saddle property, the objective function is also required to have no spurious local minima.

In summary, the geometric landscape of an objective function relates to questions concerning the existence of spurious local minima and the strict saddle property. In particular, if the function has no spurious local minima and obeys the strict saddle property, then a number of iterative algorithms such as the ones listed in \Cref{thm:global convergence} converge to a global minimum. Our goal in the next section is to show that the objective function in training a shallow linear network with a regularizer satisfies these conditions.

\section{Global Optimality in Shallow Linear Networks}
\label{sec:main results}

In this paper, we consider the following optimization problem concerning the training of a shallow linear network:
\begin{align}
\min_{\substack{\mW_1\in\R^{d_1\times d_0} \\\mW_2\in\R^{d_2\times d_1}}} f(\mW_1,\mW_2) = \frac{1}{2}\|\mW_2\mW_1\mX - \mY\|_F^2,
\label{eq:shallow linear network}\end{align}
where $\mX\in\R^{d_0\times N}$ and $\mY\in\R^{d_2\times N}$ are the input and output training examples, and $\mW_1\in\R^{d_1\times d_0}$ and $\mW_2\in\R^{d_2\times d_1}$ are the model parameters (or weights) corresponding to the first and second layers,  respectively. Throughout, we call $d_0$, $d_1$, and $d_2$ the sizes of the input layer, hidden layer, and output layer, respectively. The goal of training a neural network is to optimize the parameters $\mW_1$ and $\mW_2$ such that the output $\mW_2\mW_1\mX$ matches the desired output $\mY$.

Instead of proposing new algorithms to minimize the objective function in~\eqref{eq:shallow linear network}, we are interested in characterizing its geometric landscape by understanding the behavior of all of its critical points.

\subsection{Main results}

We present our main { theorems} concerning the behavior of all of the critical points of problem~\eqref{eq:shallow linear network}. { First,} the following result shows that the objective function of~\eqref{eq:shallow linear network} with an additional regularizer (see \eqref{low-rank for network 2}) has no spurious local minima and obeys the strict saddle property without requiring any of the following conditions that appear in { certain works discussed in} Section~\ref{sec:connectprev}: that $\mY$ is of full row rank, that $d_2 \leq d_0$, that $\mY\mX^\T (\mX\mX^\T)^{-1}\mX\mY^\T$ has $d_2$ distinct eigenvalues, that $d_1 \leq \min(d_0, d_2)$, that~\eqref{eq:cond low-rank for network} holds, or that~\eqref{eq:rank constraint} holds.

\begin{thm} Assume that $\mX\mX^\T$ is of full row rank. Then for any $\mu>0$, the following objective function
	\e\begin{split}
		g(\mW_1,\mW_2) =& \frac{1}{2}\|\mW_2\mW_1\mX - \mY\|_F^2 + \frac{\mu}{4} \rho(\mW_1,\mW_2)
	\end{split}
	\label{low-rank for network 2}\ee
	with
	\e
	\rho(\mW_1,\mW_2) : = \|\mW_2^\T \mW_2 - \mW_1 \mX \mX^\T \mW_1^\T\|_F^2
	\label{eq:rho}\ee
	obeys the following properties:
	\begin{enumerate}[label=(\roman*)]
		\item
		$g(\mW_1,\mW_2)$ has the same global minimum value as $f(\mW_1,\mW_2)$ in~\eqref{eq:shallow linear network};
		\item  any critical point $(\mW_1,\mW_2)$ of $g$ is also a critical point of $f$;
		\item $g(\mW_1,\mW_2)$ has no spurious local minima and the Hessian at any saddle point has a strictly negative eigenvalue.
	\end{enumerate}
	\label{thm:low-rank for network 2}\end{thm}

The proof of Theorem~\ref{thm:low-rank for network 2} is given in Section~\ref{sec:prf thmlow-rank for network 2}. The main idea in proving Theorem~\ref{thm:low-rank for network 2} is to connect $g(\mW_1,\mW_2)$ in~\eqref{low-rank for network 2} with the following low rank factorization problem
\[
\min_{\widetilde\mW_1,\mW_2}\frac{1}{2}\|\mW_2\widetilde\mW_1 - \widetilde\mY\|_F^2 + \frac{\mu}{4} \|\mW_2^\T \mW_2 - \widetilde\mW_1 \widetilde\mW_1^\T\|_F^2,
\]
where $\widetilde \mW_1$ and $\widetilde\mY$ are related to $\mW$ and $\mY$; see \eqref{eq:set of critical points} in Section~\ref{sec:prf thmlow-rank for network 2} for the formal definitions.

\Cref{thm:low-rank for network 2}($i$) states that the regularizer $\rho(\mW_1,\mW_2)$ in~\eqref{eq:rho} has no effect on the global minimum of the original problem, i.e., the one without this regularizer. Moreover, as established in \Cref{thm:low-rank for network 2}($ii$), any critical point of $g$ in \eqref{low-rank for network 2} is also a critical point of $f$ in \eqref{eq:shallow linear network}, but the converse is not true. With  the regularizer $\rho(\mW_1,\mW_2)$, which mostly plays the role of shrinking the set of critical points, we prove that $g$ has no spurious local minima and obeys the strict saddle property.

{
As our results hold for any $\mu>0$ and $g = f$ when $\mu = 0$, one may conjecture that these properties also hold for the original objective function $f$ under the same assumptions, i.e., assuming only that $\mX\mX^\T$ has full row rank.  This is indeed true and is formally established in the following result.
\begin{thm} Assume that $\mX$ is of full row rank. Then, the objective function $f$ appearing in~\eqref{eq:shallow linear network} has no spurious local minima and obeys the strict saddle property.
\label{thm:low-rank for network no reg}\end{thm}
The proof of \Cref{thm:low-rank for network no reg} is given in \Cref{sec:prf low-rank for network no reg}. \Cref{thm:low-rank for network no reg} builds heavily on \Cref{thm:low-rank for network 2} and on \cite[Theorem 8]{nouiehed2018learning}, which is also presented in \Cref{lem:degenerate case}. Specifically, as we have noted, \cite[Theorem 8]{nouiehed2018learning} characterizes the behavior of degenerate critical points. Using \Cref{thm:low-rank for network 2}, we further prove that any {\em non-degenerate} critical point of $f$ is either a global minimum or a strict saddle.
}

\subsection{Connection to previous work on shallow linear neural networks}
\label{sec:connectprev}

As summarized in Table~\ref{table:summary}, the results in~\cite{baldi1989neural,lu2017depth,laurent2017deep}  on characterizing the geometric landscape of the loss function in training shallow linear neural networks only consider the behavior of local minima, but not saddle points. The strict saddle property is proved only in \cite{kawaguchi2016deep} {and partly in \cite{nouiehed2018learning}}.  We first review the result in \cite{kawaguchi2016deep} concerning the optimization geometry of problem~\eqref{eq:shallow linear network}.
\begin{thm}\cite[Theorem 2.3]{kawaguchi2016deep} Assume that $\mX$ and $\mY$ are of full row rank with $d_2 \leq d_0$ and $\mY\mX^\T (\mX\mX^\T)^{-1}\mX\mY^\T$ has $d_2$ distinct eigenvalues. Then, the objective function $f$ appearing in~\eqref{eq:shallow linear network} has no spurious local minima and obeys the strict saddle property.
	\label{thm:kawaguchi2016deep}\end{thm}

Theorem~\ref{thm:kawaguchi2016deep} implies that the objective function in~\eqref{eq:shallow linear network} has benign geometric properties if $d_2 \leq d_0$ and the training samples are such that $\mX$ and $\mY$ are of full row rank and $\mY\mX^\T (\mX\mX^\T)^{-1}\mX\mY^\T$ has $d_2$ distinct eigenvalues. The recent work~\cite{lu2017depth} generalizes the first point of Theorem~\ref{thm:kawaguchi2016deep} (i.e., no spurious local minima) by getting rid of the assumption that $\mY\mX^\T (\mX\mX^\T)^{-1}\mX\mY^\T$ has $d_2$ distinct eigenvalues. However, the geometry of the saddle points is not characterized in~\cite{lu2017depth}. In~\cite{laurent2017deep}, the authors also show that the condition on $\mY\mX^\T (\mX\mX^\T)^{-1}\mX\mY^\T$ is not necessary. In particular, when applied to \eqref{eq:shallow linear network}, the result in~\cite{laurent2017deep} implies that the objective function in \eqref{eq:shallow linear network} has no spurious local minima when $d_1\leq \min(d_0,d_2)$. This condition requires that the hidden layer is narrower than the input and output layers. Again, the optimization geometry around saddle points is not discussed in \cite{laurent2017deep}.

{ We now review the more recent result in~\cite[Theorem 8]{nouiehed2018learning}.
\begin{thm}\cite[Theorem 8]{nouiehed2018learning}
The objective function $f$ appearing in~\eqref{eq:shallow linear network} has no spurious local minima. Moreover, any critical point $\mZ$ of $f$ that is degenerate (i.e., for which $\rank(\mW_2\mW_1)<\min\{d_2,d_1,d_0\}$) is either a global minimum of $f$ or a strict saddle.\label{lem:degenerate case}\end{thm}

In cases where the global minimum of $f$ is non-degenerate---for example when $\mY = \mW_2^\star\mW_1^\star\mX$ for some $\mW_2^\star$ and $\mW_1^\star$ such that $\mW_2^\star\mW_1^\star$ is non-degenerate---\Cref{lem:degenerate case} implies that all degenerate critical points are strict saddles. However, we note that the behavior of non-degenerate critical points in these cases is more important from the algorithmic point of view, since one can always check the rank of a convergent point and perturb it if it is degenerate, but this is not possible at non-degenerate convergent points. Our \Cref{thm:low-rank for network no reg} generalizes \Cref{lem:degenerate case} to ensure that every critical point that is not a global minimum is a strict saddle, regardless of its rank.
}

Next, as a direct consequence of \cite[Theorem 3]{zhu2017GlobalOptimality}, the following result also establishes certain conditions under which the objective function in \eqref{eq:shallow linear network} { with an additional regularizer (see~\eqref{low-rank for network})} has no spurious local minima and obeys the strict saddle property.
\begin{cor}\cite[Theorem 3]{zhu2017GlobalOptimality} Suppose $d_1 \leq \min(d_0, d_2)$. Furthermore, for any $d_2\times d_0$ matrix $\mA$ with $\rank(\mA)\leq 4d_1$, suppose the following holds
	\begin{align}
	\alpha \|\mA\|_F^2 \leq \trace(\mA\mX\mX^\T\mA^\T) \leq \beta \|\mA\|_F^2
	\label{eq:cond low-rank for network}\end{align}
	for some positive $\alpha$ and $\beta$ such that $\frac{\beta}{\alpha}\leq 1.5$. Furthermore, suppose $min_{\mW\in\R^{d_2\times d_0}}\|\mW\mX - \mY\|_F^2$ admits a solution $\mW^\star$ which satisfies
	\e
	0<\rank(\mW^\star) = r^\star \leq d_1.
	\label{eq:rank constraint}\ee
	Then for any $0<\mu\leq \frac{\alpha}{16}$, the following objective function
	\e\begin{split}
		h(\mW_1,\mW_2) = &\frac{1}{2}\|\mW_2\mW_1\mX - \mY\|_F^2\\& + \frac{\mu}{4} \|\mW_2^\T \mW_2 - \mW_1 \mW_1^\T\|_F^2,
	\end{split}
	\label{low-rank for network}\ee
	has no spurious local minima and the Hessian at any saddle point has a strictly negative eigenvalue with
	\begin{equation}\begin{split}
	&\lambda_{\min}\left(\nabla^2 h(\mW_1,\mW_2)\right)\leq \\
	&\left\{\begin{matrix} -0.08\alpha\sigma_{d_1}(\mW^\star), & d_1 = r^\star \\
	-0.05\alpha\cdot \min\left\{\sigma_{r_c}^2(\mW_2\mW_1),\sigma_{r^\star}(\mW^\star)\right\}, & d_1>r^\star\\
	-0.1\alpha \sigma_{r^\star}(\mW^\star), & r_c = 0,
	\end{matrix}\right.
	\label{eq:strict saddle for h}\end{split}\end{equation}
	where $r_c\leq d_1$ is the rank of $\mW_1\mW_2$, $\lambda_{\min}(\cdot)$ represents the smallest eigenvalue, and $\sigma_\ell(\cdot)$ denotes the $\ell$-th largest singular value.
	\label{thm:low-rank for network}\end{cor}
Corollary~\ref{thm:low-rank for network}, following from \cite[Theorem 3]{zhu2017GlobalOptimality}, utilizes a regularizer $\|\mW_2^\T \mW_2 - \mW_1 \mW_1^\T\|_F^2$ which balances the energy between $\mW_1$ and $\mW_2$ and has the effect of shrinking the set of critical points. This allows one to show that each critical point is either a global minimum or a strict saddle. Similar to \Cref{thm:low-rank for network 2}($i$), this regularizer also has no effect on the global minimum of the original problem~\eqref{eq:shallow linear network}. 

As we explained before, \Cref{thm:kawaguchi2016deep} implicitly requires that $\min(d_0,d_1)\geq d_2$ and $\rank(\mW_2^\star\mW_1^\star)\geq d_2$. On the other hand, \Cref{thm:low-rank for network} requires $d_1\leq \min(d_0,d_2)$ and \eqref{eq:cond low-rank for network}. When $d_1\leq \min(d_0,d_2)$, the hidden layer is narrower than the input and output layers. Note that~\eqref{eq:cond low-rank for network} has nothing to do with the underlying network parameters $\mW_1^\star$ and $\mW_2^\star$, but requires the training data matrix $\mX$ to act as an isometry operator for rank-$4d_1$ matrices. To see this, we rewrite
\[
\trace(\mA\mX\mX^\T\mA^\T) = \sum_{i=1}^N \vx_i^\T\mA^\T\mA\vx_i = \sum_{i=1}^N \langle \vx_i\vx_i^\T,\mA^\T\mA\rangle
\]
which is a sum of the rank-one measurements of $\mA^\T\mA$.

Unlike \Cref{thm:kawaguchi2016deep}, which requires that $\mY\mY^\T$ is of full rank and $d_2 \leq d_0$, and unlike \Cref{thm:low-rank for network}, which requires \eqref{eq:cond low-rank for network} and $d_1\leq\min(d_0,d_2)$, \Cref{thm:low-rank for network 2} {and \Cref{thm:low-rank for network no reg}} only necessitate that $\mX\mX^\T$ is full rank and have no condition on the size of $d_0$, $d_1$, and $d_2$. As we explained before, suppose $\mY$ is generated as $\mY = \mW_2^\star\mW_1^\star\mX$, where $\mW_2^\star$ and $\mW_1^\star$ are the underlying weights to be recovered. Then the full-rank assumption of $\mY\mY^{\T} = \mW_2^\star\mW_1^\star\mX\mX^\T \mW_1^{\star\T}\mW_2^{\star\T}$ at least requires $\min(d_0,d_1)\geq d_2$ and $\rank(\mW_2^\star\mW_1^\star)\geq d_2$. In other words, \Cref{thm:kawaguchi2016deep} necessitates that the hidden layer is wider than the output, while \Cref{thm:low-rank for network 2} {and \Cref{thm:low-rank for network no reg}} work for networks where the hidden layer is narrower than the input and output layers. On the other hand, \Cref{thm:low-rank for network 2} {and \Cref{thm:low-rank for network no reg}} allow for the hidden layer of the network to be either narrower or wider than the input and the output layers.

Finally, consider a three-layer network with $\mX = \mId$. In this case, \eqref{eq:shallow linear network} reduces to a matrix factorization problem where $f(\mW_1,\mW_2) = \|\mW_2\mW_1 - \mY\|_F^2$ and the regularizer in~\eqref{low-rank for network 2} is the same as the one in~\eqref{low-rank for network}. Theorem~\ref{thm:kawaguchi2016deep} requires that $\mY$ is of full row rank and has $d_2$ distinct singular values. For the matrix factorization problem, we know from \Cref{thm:low-rank for network} that for any $\mY$, $h$ has benign geometry (i.e., no spurious local minima and the strict saddle property) as long as $d_1 \leq \min(d_0,d_2)$. As a direct consequence of \Cref{thm:low-rank for network 2} {and \Cref{thm:low-rank for network no reg}}, this benign geometry is also preserved even when $d_1> d_0$ or $d_1 > d_2$ for matrix factorization via minimizing
\[
g(\mW_1,\mW_2) = \|\mW_2\mW_1 - \mY\|_F^2 + \frac{\mu}{4}\|\mW_2^\T \mW_2 - \mW_1 \mW_1^\T\|_F^2,
\]
{where $\mu\ge 0$ (note that one can get rid of the regularizer by setting $\mu = 0$).}

\section{Proof of Main Results}
\label{sec:prf}
\subsection{Proof of \Cref{thm:low-rank for network 2}}\label{sec:prf thmlow-rank for network 2}
In this section, we prove \Cref{thm:low-rank for network 2}. We first show that the regularizer\ in~\eqref{low-rank for network 2} has no effect on the global minimum of the original problem.  It is clear that $g(\mW_1,\mW_2)\geq f(\mW_1,\mW_2)$ for any $\mW_1,\mW_2$, where we repeat that
\[
f(\mW_1,\mW_2) = \frac{1}{2}\|\mW_2\mW_1\mX - \mY\|_F^2.
\]
Suppose the row rank of $\mX$ is $d_0'\le d_0$. Let
\e
\mX = \mU\mSigma\mV^\T
\label{eq:SVD X}\ee
be a reduced SVD of $\mX$, where $\mSigma$ is a $d_0'\times d_0'$ diagonal matrix with positive diagonals. Then,
\begin{align*}
f(\mW_1,\mW_2) &= \frac{1}{2}\|\mW_2\mW_1\mU\mSigma - \mY\mV\|_F^2 + \|\mY\|_F^2 - \|\mY\mV\|_F^2\\ &= f_1(\mW_1,\mW_2) + C,
\end{align*}
where $f_1(\mW_1,\mW_2) = \frac{1}{2}\|\mW_2\mW_1\mU\mSigma - \mY\mV\|_F^2$ and $C = \|\mY\|_F^2 - \|\mY\mV\|_F^2$.
Denote by $(\mW_1^\star,\mW_2^\star)$  a global minimum of  $f_1(\mW_1,\mW_2)$ :
\[
(\mW_1^\star,\mW_2^\star) = \argmin_{\mW_1,\mW_2} f(\mW_1,\mW_2) = \argmin_{\mW_1,\mW_2} f_1(\mW_1,\mW_2).
\]

Let $\mW_2^\star\mW_1^\star\mU\mSigma = \mP_1\mOmega\mQ_1^\T$ be  a reduced SVD of $\mW_2^\star\mW_1^\star \mU\mSigma$, where $\mOmega$ is a diagonal matrix with positive diagonals. Let $\widehat\mW_2 = \mP_1\mOmega^{1/2}$ and  $\widehat\mW_1 = \mOmega^{1/2}\mQ_1^\T\mSigma^{-1}\mU^\T$. It follows that
\begin{align*}
&\widehat\mW_2^\T \widehat\mW_2 - \widehat\mW_1 \mX\mX^\T\widehat\mW_1^\T =\mOmega - \mOmega = \mzero,\\ &\widehat\mW_2 \widehat\mW_1 \mU\mSigma = \mP_1\mOmega\mQ_1^\T = \mW_2^\star\mW_1^\star\mU\mSigma,
\end{align*}
which implies that $f_1(\mW_1^\star,\mW_2^\star) = f_1(\widehat\mW_1,\widehat\mW_2)$ and
\begin{align*}
&f(\mW_1^\star,\mW_2^\star) = f_1(\mW_1^\star,\mW_2^\star) + C = f_1(\widehat\mW_1,\widehat\mW_2) + C\\& = f(\widehat\mW_1,\widehat\mW_2) = g(\widehat\mW_1,\widehat\mW_2)
\end{align*}
since $\|\widehat\mW_2^\T \widehat\mW_2 - \widehat\mW_1 \mX\mX^\T\widehat\mW_1^\T\|_F^2= 0$. This further indicates that $g$ and $f$ have the same global optimum (since $g(\mW_1,\mW_2)\geq f(\mW_1,\mW_2)$ for any $(\mW_1,\mW_2)$).

In the rest of the proof of Theorem~\ref{thm:low-rank for network 2} we characterize the behavior of all the critical points of the objective function $g$ in~\eqref{low-rank for network 2}. In particular, we show that any critical point of $g$ is also a critical point of $f$, and if it is not a global minimum of~\eqref{low-rank for network 2}, then it is a strict saddle, i.e., its Hessian has at least one negative eigenvalue.

To that end, we first establish the following result that characterizes all the critical points of $g$.
\begin{lem}\label{lem:set of critical points}
	Let $\mX = \mU\mSigma\mV^\T$ be an SVD of $\mX$ as in \eqref{eq:SVD X}, where $\mSigma$ is a diagonal matrix with positive diagonals $\sigma_1\geq \sigma_2 \geq \cdots \geq \sigma_{d_0}>0$.
	Let $\widetilde \mY := \mY\mV =\mP \mLambda \mQ^\T = \sum_{j=1}^r\lambda_j \vp_j\vq_j^\T$ be a reduced SVD of $\widetilde \mY$, where $r$ is the rank of $\widetilde \mY$. Then any critical point $\mZ = \begin{bmatrix} \mW_2 \\ \mW_1^\T \end{bmatrix}$ of \eqref{low-rank for network 2}  satisfies
	\e
	\mW_2^\T\mW_2 = \mW_1\mX\mX^\T\mW_1^\T.
	\label{eq:W2W2 = W1X W1X}\ee
	Furthermore, any $\mZ = \begin{bmatrix} \mW_2 \\ \mW_1^\T \end{bmatrix}$ is a critical point of $g(\mZ)$ if and only if $\mZ\in\calC_g$ with
	\e\begin{split}
		\calC_g = \bigg\{\mZ &= \begin{bmatrix}\widetilde \mW_2 \mR^\T \\ \mU \mSigma^{-1}\widetilde \mW_1^\T\mR^\T \end{bmatrix}:\widetilde\mZ = \begin{bmatrix} \widetilde\mW_2 \\ \widetilde\mW_1^\T \end{bmatrix},\\ &\widetilde\vz_i\in \left\{\sqrt{\lambda_1}\begin{bmatrix}\vp_1\\\vq_1\end{bmatrix},\ldots,\sqrt{\lambda_r}\bmat\vp_r\\ \vq_r\emat,\vzero  \right\},\\ & \widetilde\vz_i^\T\widetilde\vz_j = 0, \ \forall i\neq j,\mR\in\setO_{d_1}\bigg\},
	\end{split}\label{eq:set of critical points}\ee
	where $\widetilde \vz_i$ denotes the $i$-th column of $\widetilde \mZ$.
\end{lem}
The proof of \Cref{lem:set of critical points} is in Appendix~\ref{sec:prf lem set of critical points}.
From~\eqref{eq:W2W2 = W1X W1X}, $g(\mZ) = f(\mZ)$ at any critical point $\mZ$. We compute the gradient of the regularizer $\rho(\mW_1,\mW_2)$ as
\begin{align*}
&\nabla_{\mW_1}\rho(\mW_1,\mW_2)\\&: =-\mu\mW_2(\mW_2^\T\mW_2- \mW_1\mX\mX^\T\mW_1^\T)\mW_1\mX\mX^\T,\\
&\nabla_{\mW_2}\rho(\mW_1,\mW_2)\\&: =\mu\mW_2(\mW_2^\T\mW_2- \mW_2\mX\mX^\T\mW_1^\T)\mW_1\mX\mX^\T.
\end{align*}
Plugging~\eqref{eq:W2W2 = W1X W1X} into the above equations gives
\[
\nabla_{\mW_1}\rho(\mW_1,\mW_2) = \nabla_{\mW_2}\rho(\mW_1,\mW_2) = \mzero
\]
for any critical point $(\mW_1,\mW_2)$ of $g$. This further implies that if $\mZ$ is a critical point of $g$, then it must also be a critical point of of $f$ since $\nabla g(\mZ) = \nabla f(\mZ) + \nabla \rho(\mZ) $ and both $\nabla g(\mZ)  = \nabla \rho(\mZ) = \vzero$, so that
\e
\nabla f(\mZ) = \vzero.
\label{eq:critical point of f} \ee
This proves \Cref{thm:low-rank for network 2}($ii$).

To further classify the critical points into categories such as local minima and saddle points, for any $\mZ\in\setC$, we compute the objective value at this point as
\begin{align*}
g(\mZ) &= \frac{1}{2}\|\mW_2\mW_1\mX - \mY\|_F^2 = \frac{1}{2}\|\widetilde\mW_2\widetilde\mW_1\mSigma^{-1}\mU^\T\mX - \mY\|_F^2\\
& = \frac{1}{2}\|\widetilde\mW_2\widetilde\mW_1 - \mY\mV\|_F^2 + \|\mY\|_F^2 - \|\mY\mV\|_F^2,
\end{align*}
where $\mU\mSigma\mV^\T$ is a reduced SVD of $\mX$ as defined in \eqref{eq:SVD X}, and $\widetilde \mW_2$ and $\widetilde \mW_1$ are defined in \eqref{eq:set of critical points}.
Noting that $\|\mY\|_F^2 - \|\mY\mV\|_F^2$ is a constant in terms of the variables $\mW_2$ and $\mW_1$, we conclude that $\mZ$ is a global minimum of $g(\mZ)$ if and only if  $\widetilde \mZ$ is a global minimum of
\e
\widetilde g(\widetilde \mZ) :=\frac{1}{2}\|\widetilde\mW_2\widetilde\mW_1 - \mY\mV\|_F^2.
\label{eq:low rank to YV}\ee

\begin{lem} With the same setup as in \Cref{lem:set of critical points}, let $\setC$ be defined in \eqref{eq:set of critical points}. Then all local minima of \eqref{low-rank for network 2} belong to the following set (which contains all the global solutions of \eqref{low-rank for network 2})
	\e\begin{split}
		\calX_g = \bigg\{\mZ &=  \begin{bmatrix}\widetilde \mW_2 \mR \\ \mU \mSigma^{-1}\widetilde \mW_1^\T\mR \end{bmatrix} \in \setC: \|\widetilde\mW_2\widetilde\mW_1 - \mY\mV\|_F^2 \\&\qquad= \min_{\mA\in\R^{d_2\times d_1},\mB\in\R^{d_1\times d_0}} \|\mA\mB - \mY\mV\|_F^2\bigg\}.
	\end{split}\label{eq:set of minima}\ee
	Any $\mZ\in\setC_g\setminus\setX_g$ is a strict saddle of $g(\mZ)$ satisfying:
	\begin{itemize}[label=\tiny\textbullet]
		\item if $r\leq d_1$, then
		\begin{align}
		\lambda_{\min}(\nabla^2g(\mW))\leq -2\frac{\lambda_r}{1 + \sum_i \sigma_i^{-2}};
		\label{eq:strict saddle 1}\end{align}
		\item if $r> d_1$, then
		\begin{align}
		\lambda_{\min}(\nabla^2g(\mW))\leq -2\frac{\lambda_{d_1} - \lambda_{r'}}{1 + \sum_i \sigma_i^{-2}},
		\label{eq:strict saddle 2}\end{align}
		where $\lambda_{r'}$ is the largest singular value of $\widetilde \mY$ that is strictly smaller than $\lambda_{d_1}$.
	\end{itemize}
	\label{lem:local minimum}\end{lem}
The proof of \Cref{lem:local minimum} is given in Appendix~\ref{sec:prf lem local minimum}.  \Cref{lem:local minimum} states that any critical point of $g$ is either a global minimum or a strict saddle. This proves \Cref{thm:low-rank for network 2}($iii$) and thus we complete the proof of \Cref{thm:low-rank for network 2}.

\subsection{Proof of \Cref{thm:low-rank for network no reg}}
\label{sec:prf low-rank for network no reg}
{
Building on  \Cref{thm:low-rank for network 2} and \Cref{lem:degenerate case}, we now consider the landscape of $f$ in \eqref{eq:shallow linear network}.  Let $\calC_f$ denote the set of critical points of $f$:
\[
\calC_f = \left\{\mZ: \nabla f(\mZ)= \vzero  \right\}.
\]
Our goal is to characterize the behavior of all critical points that are not global minima. In particular, we want to show that every critical point of $f$ is either a global minimum or a strict saddle.

Let  $\mZ = \begin{bmatrix} \mW_2 \\ \mW_1^\T \end{bmatrix}$ be any critical point in $\calC_f$. According to \Cref{lem:degenerate case}, when $\mW_2\mW_1$ is degenerate (i.e., $\rank(\mW_2\mW_1) < \min\{d_2,d_1,d_0\}$), then $\mZ$ must be either a global minimum or a strict saddle. We now assume the other case that $\mW_2\mW_1$ is non-degenerate.

Let $\mW_2\mW_1\mU\mSigma = \mPhi \mTheta \mPsi^\T$ be a reduced SVD of $\mW_2\mW_1\mU\mSigma$, where $\mTheta$ is a diagonal and square matrix with positive singular values, and $\mPhi$ and $\mPsi$ are orthonormal matrices of proper dimension.  We now construct
\e\begin{split}
	&\overline\mW_2 = \mW_2\mW_1\mU\mSigma\mPsi \mTheta^{-1/2} = \mPhi\mTheta^{1/2},\\
	&\overline\mW_1 = \mTheta^{-1/2}\mPhi^\T \mW_2\mW_1 = \mTheta^{1/2}\mPsi^\T\mSigma^{-1}\mU^\T.
\end{split}\ee
The above constructed pair $\overline\mZ = \begin{bmatrix} \overline\mW_2 \\ \overline\mW_1^\T \end{bmatrix}$ satisfies
\e\begin{split}
\overline\mW_2^\T \overline\mW_2 = \overline\mW_1\mX\mX^\T \overline\mW_1^\T,~ \overline\mW_2\overline\mW_1 =  \mW_2\mW_1.
\end{split}
\label{eq:Zbar balanced}\ee

Note that here $\overline\mW_2$ (resp. $\overline\mW_1$) have different numbers of columns (resp. rows) than $\mW_2$ (resp. $\mW_1$). We denote by
\[
\overline f(\overline \mW_1, \overline \mW_2) = \frac{1}{2}\|\overline\mW_2\overline\mW_1 \mU\mSigma- \mY\mV\|_F^2.
\]

Since $\mZ\in\calC_f$, we have $\nabla_{\mW_1}f(\mW_1,\mW_2) = \mzero$ and $\nabla_{\mW_2}f(\mW_1,\mW_2) =\mzero$. It follows that
\begin{align*}
\nabla_{\overline\mW_2} \overline f(\overline \mW_1, \overline \mW_2) &= (\overline\mW_2\overline\mW_1 \mU\mSigma- \mY\mV)(\overline\mW_1 \mU\mSigma)^\T\\
&= \nabla_{\mW_2} f(\mW_1,  \mW_2)\mW_2^\T\mPhi\mTheta^{-1/2}\\ &= \vzero.
\end{align*}
And similarily, we have
\begin{align*}
&\nabla_{\overline\mW_1} \overline f(\overline \mW_1, \overline \mW_2)\\&=\mTheta^{-1/2} \mPsi^\T\mSigma^{-1}\mU^\T\mW_1^\T \nabla_{\mW_1} f(\mW_1,  \mW_2)= \vzero,
\end{align*}
which together with the above inequation implies that $\overline\mZ$ is also a critical point  of $\overline f$. Due to \eqref{eq:Zbar balanced} which states that $\overline\mZ$ also satisfies \eqref{eq:W2W2 = W1X W1X},  it follows from the same arguments used in \Cref{lem:set of critical points} and \Cref{lem:local minimum}  that $\overline\mZ$ is either a global minimum or a strict saddle of $\overline f$. Moreover, since $\overline\mW_2\overline\mW_1$ has the same rank as $\mW_2\mW_1$ which is assumed to be non-degenerate, we have that $\overline\mZ$ is a global minimum of $\overline f$ if and only if
\[
\|\overline\mW_2\overline\mW_1 \mU\mSigma- \mY\mV\|_F^2 = \min_{\substack{\mA\in\R^{d_2\times d_1}\\\mB\in\R^{d_1\times d_0}}} \|\mA\mB - \mY\mV\|_F^2,
\]
where the minimum of the right hand side is also achieved by the global minimum of $f$ according to \eqref{eq:set of minima}. Therefore, if $\overline\mZ$ is a global minimum of $\overline f$, then $\mZ$ is also a global minimum of $f$.

Now we consider the other case when $\overline\mZ$ is not a global minimum of $\overline f$,  i.e., it is a strict saddle. In this case, there exists $\overline\mDelta = \begin{bmatrix}\overline\mDelta_2\\\overline\mDelta_1^\T\end{bmatrix}$ such that
\[
[\nabla^2 \overline f(\overline\mW_1,\overline\mW_2)](\overline\mDelta,\overline\mDelta) <0.
\]
Now construct
\begin{align*}
&\mDelta_{1} = \mW_1\mU\mSigma\mPsi \mTheta^{-1/2}\overline \mDelta_1\\
&\mDelta_{2} = \overline\mDelta_{2}\mTheta^{-1/2}\mPhi^\T\mW_2
\end{align*}
which satisfies
\begin{align*}
&\mW_2\mDelta_1 = \overline\mW_2 \overline \mW_1, \ \mDelta_{\mW_2}\mW_1 = \overline \mDelta_2 \overline \mW_1, \ \mDelta_2\mDelta_1 = \overline\mDelta_2 \overline\mDelta_1.
\end{align*}

By the Hessian quadratic form given in \eqref{eq:Hessian} (ignoring the $\mu$ terms), we have
\begin{align*}
[\nabla^2 f(\mZ)](\mDelta,\mDelta) = [\nabla^2 \overline f(\overline\mZ)](\overline\mDelta,\overline\mDelta) <0,
\end{align*}
which implies that $\mZ$ is a strict saddle  of $f$. This completes the proof of \Cref{thm:low-rank for network no reg}.

}

\section{Conclusion}
We consider the optimization landscape of the objective function in training shallow linear networks. In particular,  we proved that the corresponding optimization problems under a very mild condition have a simple landscape: there are no spurious local minima and any critical point is either a local (and thus also global) minimum or a strict saddle such that the Hessian evaluated at this point has a strictly negative eigenvalue. These properties guarantee that a number of iterative optimization algorithms (especially gradient descent, which is widely used in training neural networks) converge to a global minimum from either a random initialization  or an arbitrary initialization depending on the specific algorithm used. {It would be of interest to prove similar geometric properties for the training problem without the mild condition on the row rank of $\mX$.}
\appendix

\section{Proof of \Cref{lem:set of critical points}}\label{sec:prf lem set of critical points}
We first prove the direction $\Rightarrow$.
Any critical point $\mZ$ of $g(\mZ)$ satisfies  $\nabla g(\mZ)=\vzero$, i.e.,
\e\begin{split}
	&\nabla_{\mW_1} g(\mW_1,\mW_2)= \mW_2^\T(\mW_2 \mW_1\mX - \mY)\mX^\T\\& - \mu(\mW_2^\T\mW_2 - \mW_1\mX\mX^\T\mW_1^\T)\mW_1\mX\mX^\T = \mzero,\label{eq:grad W1}
\end{split}\ee
and
\e\begin{split}
	&\nabla_{\mW_2} g(\mW_1,\mW_2)= (\mW_2 \mW_1\mX - \mY)\mX^\T\mW_1^\T\\ &+ \mu\mW_2(\mW_2^\T\mW_2 - \mW_1\mX\mX^\T\mW_1^\T) = \mzero.\label{eq:grad W2}
\end{split}\ee
By \eqref{eq:grad W1}, we obtain
\e\begin{split}
	&\mW_2^\T(\mW_2 \mW_1\mX - \mY)\mX^\T\\& = \mu(\mW_2^\T\mW_2 - \mW_1\mX\mX^\T\mW_1^\T)\mW_1\mX\mX^\T.
\end{split}
\label{eq:grad W1 0 gives}\ee
Multiplying \eqref{eq:grad W2} on the left by $\mW_2^\T$ and plugging the result with the expression for $\mW_2^\T(\mW_2 \mW_1\mX - \mY)\mX^\T$ in \eqref{eq:grad W1 0 gives} gives
\begin{align*}
&(\mW_2^\T\mW_2 - \mW_1\mX\mX^\T\mW_1^\T)\mW_1\mX\mX^\T\mW_1^\T\\& + \mW_2^\T \mW_2(\mW_2^\T\mW_2 - \mW_1\mX\mX^\T\mW_1^\T) = \vzero,
\end{align*}
which is equivalent to
\[
\mW_2^\T \mW_2\mW_2^\T\mW_2 = \mW_1\mX\mX^\T\mW_1^\T\mW_1\mX\mX^\T\mW_1^\T.
\]

Note that $\mW_2^\T\mW_2$ and $\mW_1\mX\mX^\T\mW_1^\T$ are the principal square roots (i.e., PSD square roots) of $\mW_2^\T \mW_2\mW_2^\T\mW_2$ and $\mW_1\mX\mX^\T\mW_1^\T\mW_1\mX\mX^\T\mW_1^\T$, respectively. Utilizing the result that a PSD matrix $\mA$ has a unique PSD matrix $\mB$ such that $\mB^k = \mA$ for any $k\geq 1$~\cite[Theorem 7.2.6]{horn2012matrix}, we obtain
\[
\mW_2^\T\mW_2 = \mW_1\mX\mX^\T\mW_1^\T
\]
for any critical point $\mZ$. Plugging into~\eqref{eq:grad W1} and \eqref{eq:grad W2}, any critical point $\mZ$ of $g(\mZ)$ satisfies
\e\begin{split}
	&\mW_2^\T(\mW_2 \mW_1\mX - \mY)\mX^\T  = \mzero,\\
	&(\mW_2 \mW_1\mX - \mY)\mX^\T\mW_1^\T = \mzero.\label{eq:critical point no reg}
\end{split}\ee

To show \eqref{eq:set of critical points}, let $\mW_2 = \mL\mPi\mR^\T$ be a full  SVD of $\mW_2$, where $\mL\in\R^{d_2\times d_2}$ and $\mR\in\R^{d_1\times d_1}$ are orthonormal matrices. Define
\e
\widetilde \mW_2 = \mW_2 \mR =  \mL\mPi, \ \widetilde \mW_1 = \mR^\T\mW_1\mU\mSigma.
\label{eq:W to tilde W}\ee
Since $\mW_1\mX\mX^\T\mW_1^\T = \mW_2^\T\mW_2 $ (see \eqref{eq:W2W2 = W1X W1X}), we have
\e
\widetilde\mW_1\widetilde\mW_1^\T = \widetilde\mW_2^\T \widetilde\mW_2 = \mPi^\T\mPi.
\label{eq:WT W diagnal}\ee
Noting that $\mPi^\T\mPi$ is a diagonal matrix with nonnegative diagonals, it follows that $\widetilde\mW_1^\T$ is an orthogonal matrix, but possibly includes zero columns.

Due to \eqref{eq:critical point no reg}, we have
\e\begin{split}
	&\widetilde\mW_2^\T(\widetilde\mW_2 \widetilde\mW_1 - \mY\mV)\mSigma\mU^\T\\&  = \mR^\T (\mW_2^\T(\mW_2 \mW_1\mX - \mY)\mX^\T) = \mzero,\\
	&(\widetilde\mW_2 \widetilde\mW_1 - \mY\mV)\mSigma\mU^\T\widetilde\mW_1^\T\\& = (\mW_2 \mW_1\mX - \mY)\mX^\T\mW_1^\T \mR = \mzero,\label{eq:critical point tildeW no reg}
\end{split}\ee
where we utilized the reduced SVD decomposition $\mX = \mU\mSigma\mV^\T$ in~\eqref{eq:SVD X}.
Note that the diagonals of $\mSigma$ are all positive and recall
\[
\widetilde \mY = \mY\mV.
\]
Then \eqref{eq:critical point tildeW no reg} gives
\e\begin{split}
	&\widetilde\mW_2^\T(\widetilde\mW_2 \widetilde\mW_1 - \widetilde\mY) = \mzero,\\
	&(\widetilde\mW_2 \widetilde\mW_1 - \widetilde\mY)\widetilde\mW_1^\T = \mzero.\label{eq:critical point tildeW no reg v2}
\end{split}\ee

We now compute all $\widetilde \mW_2$ and $\widetilde\mW_1$ satisfying \eqref{eq:critical point tildeW no reg v2}. To that end, let $\vphi\in\R^{d_2}$ and $\vpsi\in\R^{d_0}$ be the $i$-th column and the $i$-th row of $\widetilde \mW_2$ and $\widetilde \mW_1$, respectively. Due to \eqref{eq:WT W diagnal}, we have
\e
\|\vphi\|_2 = \|\vpsi\|_2.
\label{eq:phi psi same norm}\ee
It follows from \eqref{eq:critical point tildeW no reg v2} that
\begin{align}
\widetilde \mY^\T \vphi = \|\vphi\|_2^2 \vpsi,\label{eq:Y p q 1}\\
\widetilde \mY \vpsi = \|\vpsi\|_2^2 \vphi.\label{eq:Y p q 2}
\end{align}
Multiplying \eqref{eq:Y p q 1} by $\widetilde \mY$ and plugging \eqref{eq:Y p q 2} into the resulting equation gives
\e
\widetilde \mY \widetilde \mY^\T\vphi = \|\vphi\|_2^4\vphi,
\label{eq:Y Yt p}\ee
where we used \eqref{eq:phi psi same norm}. Similarly, we have
\e
\widetilde \mY^\T \widetilde \mY\vpsi = \|\vpsi\|_2^4\vpsi.
\label{eq:Yt Y q}\ee

Let $\widetilde \mY = \mP \mLambda \mQ^\T = \sum_{j=1}^r\lambda_j \vp_j\vq_j^\T$ be the reduced SVD of $\widetilde \mY$. It follows from \eqref{eq:Y Yt p}  that $\vphi$ is either a zero vector (i.e., $\vphi = \vzero$), or a left singular vector of $\widetilde \mY$ (i.e., $\vphi = \alpha \vp_j$ for some $j\in[r]$). Plugging $\vphi = \alpha \vp_j$ into \eqref{eq:Y Yt p} gives
\[
\lambda_j^2 = \alpha^4.
\]
Thus, $\vphi = \pm \sqrt{\lambda_j}\vp_j$. If $\vphi = \vzero$, then due to \eqref{eq:phi psi same norm}, we have $\vpsi = \vzero$. If $\vphi = \pm \sqrt{\lambda_j}\vp_j$, then plugging into \eqref{eq:Y p q 1} gives
\[
\vpsi = \pm \sqrt{\lambda_j}\vq_j.
\]
Thus, we conclude that
\[
(\vphi,\vpsi)\in\left\{\pm\sqrt{\lambda_1}(\vp_1,\vq_1),\ldots,\pm\sqrt{\lambda_r}(\vp_r,\vq_r),(\vzero,\vzero)  \right\},
\]
which together with \eqref{eq:WT W diagnal} implies that any critical point $\mZ$ belongs to \eqref{eq:set of critical points} by absorbing the sign $\pm$ into $\mR$.

We now prove the other direction $\Rightarrow$. For any $\mZ\in\setC$, we compute the gradient of $g$ at this point and directly verify it satisfies \eqref{eq:grad W1} and \eqref{eq:grad W2}, i.e., $\mZ$ is a critical point of $g(\mZ)$. This completes the proof of \Cref{lem:set of critical points}.

\section{Proof of \Cref{lem:local minimum}}
\label{sec:prf lem local minimum}
Due to the fact that $\mZ$ is a global minimum of $g(\mZ)$ if and only if  $\widetilde \mZ$ is a global minimum of $\widetilde g(\widetilde \mZ)$, we know any $\mZ\in\setX$ is a global minimum of $g(\mZ)$. The rest is to show that any $\mZ\in \setC\setminus\setX$ is a strict saddle. For this purpose, we first compute the Hessian quadrature form $\nabla^2 g(\mZ)[\mDelta,\mDelta]$ for any $\mDelta = \begin{bmatrix}\mDelta_2\\\mDelta_1^\T\end{bmatrix}$ (with $\mDelta_1\in\R^{d_1\times d_0},\mDelta_2\in\R^{d_2\times d_1}$) as
\e\begin{split}
	&\nabla^2 g(\mZ)[\mDelta,\mDelta]\\ &= \left\|(\mW_2 \mDelta_1 + \mDelta_2\mW_1)\mX \right\|_F^2\\&\ + 2\left\langle \mDelta_2\mDelta_1,(\mW_2 \mW_1 \mX - \mY)\mX^\T \right\rangle\\
	&\ + \mu\big(\langle \mW_2^\T \mW_2 - \mW_1 \mX \mX^\T \mW_1^\T,\mDelta_2^\T\mDelta_2 - \mDelta_1\mX\mX^\T\mDelta_1^\T\rangle+ \\
	& \   \frac{1}{2}\|\mW_2^\T\mDelta_2 + \mDelta_2^\T\mW_2 - \mW_1\mX\mX^\T\mDelta_1^\T - \mDelta_1\mX\mX^\T\mW_1^T\|_F^2  \big)\\
	&= \left\|(\mW_2 \mDelta_1 + \mDelta_2\mW_1)\mX_1 \right\|_F^2\\
	&\ + 2\left\langle \mDelta_2\mDelta_1,(\mW_2 \mW_1 \mX - \mY)\mX^\T \right\rangle+\\
	&\ \frac{\mu}{2}\|\mW_2^\T\mDelta_2 + \mDelta_2^\T\mW_2 - \mW_1\mX\mX^\T\mDelta_1^\T - \mDelta_1\mX\mX^\T\mW_1^T\|_F^2,
\end{split}\label{eq:Hessian}\ee
where the second equality follows because any critical point $\mZ$ satisfies \eqref{eq:W2W2 = W1X W1X}. We continue the proof by considering two cases in which we provide explicit expressions for the set $\setX$ that contains all the global minima and construct a negative direction for $g$ at all the points $\setC\setminus\setX$.

Case $i$: $r\leq d_1$. In this case, $\min \widetilde g(\widetilde \mZ) = 0$ and $\widetilde g(\widetilde \mZ)$ achieves its global minimum $0$ if and only if $\widetilde\mW_2\widetilde\mW_1 = \mY\mV$. Thus, we rewrite $\setX$ as
\e\begin{split}
	\calX = \bigg\{\mZ =  \begin{bmatrix}\widetilde \mW_2 \mR \\ \mU \mSigma^{-1}\widetilde \mW_1^\T\mR \end{bmatrix} \in \setC: \widetilde\mW_2\widetilde\mW_1 = \mY\mV\bigg\},
\end{split}\nonumber\ee
which further implies that
\begin{align*}
\setC \setminus \calX = \bigg\{\mZ =  &\begin{bmatrix}\widetilde \mW_2 \mR \\ \mU \mSigma^{-1}\widetilde \mW_1^\T\mR \end{bmatrix} \in \setC:\\& \mY\mV - \widetilde\mW_2\widetilde\mW_1 =\sum_{i\in\Omega}\lambda_i\vp_i\vq_i^\T,\Omega\subset[r] \bigg\}.
\end{align*}
Thus, for any $\mZ\in \setC \setminus \calX$, the corresponding $\widetilde\mW_2\widetilde\mW_1$ is a low-rank approximation to $\mY\mV$.

Let $k\in \Omega$. We have
\e
\vp_k^\T \widetilde \mW_2 = \vzero, \ \widetilde\mW_1 \vq_k = \vzero.
\label{eq:strict saddle k property}\ee
In words, $\vp_k$ and $\vq_k$ are orthogonal to $\widetilde \mW_2$ and $\widetilde \mW_1$, respectively. Let $\valpha\in\R^{d_1}$ be the eigenvector associated with the smallest eigenvalue of $\widetilde\mZ^\T\widetilde \mZ$. Note that such $\valpha$ simultaneously lives in the null spaces of $\widetilde \mW_2$ and $\widetilde \mW_1^\T$ since $\widetilde\mZ$ is rank deficient, indicating
\begin{align*}
0 =\valpha^\T \widetilde\mZ^\T\widetilde\mZ \valpha =  \valpha^\T \widetilde\mW_2^\T\widetilde\mW_2 \valpha +  \valpha^\T \widetilde\mW_1\widetilde\mW_1^\T \valpha, \end{align*}
which further implies
\begin{align}
\widetilde\mW_2 \valpha = \vzero,\ \widetilde\mW_1^\T \valpha = \vzero.
\label{eq:strict saddle alpha}\end{align}
With this property, we construct $\mDelta$ by setting $\mDelta_{2} = \vp_k\valpha^\T\mR$ and $\mDelta_{1} = \mR^\T\valpha\vq_k^\T \mSigma^{-1}\mU^\T$.

Now we show that $\mZ$ is a strict saddle by arguing that $g(\mZ)$ has a strictly negative curvature along the constructed direction $\mDelta$, i.e., $[\nabla^2g(\mZ)](\mDelta,\mDelta)<0$. For this purpose, we compute the three terms in \eqref{eq:Hessian} as follows:
\begin{align}
\left\|(\mW_2 \mDelta_1 + \mDelta_2\mW_1)\mX_1 \right\|_F^2 = 0
\label{eq:third term 0}\end{align}
since $\mW_2\mDelta_1 = \mW_2\mR^\T\valpha\vq_k^\T \mSigma^{-1}\mU^\T = \widetilde\mW_2\valpha\vq_k^\T \mSigma^{-1}\mU^\T = \vzero$ and $\mDelta_2\mW_1 = \vp_k\valpha^\T\mR\mW_1 = \vp_k\valpha^\T\widetilde\mW_1 = \vzero$ by utilizing \eqref{eq:strict saddle alpha};
\[
\|\mW_2^\T\mDelta_2 + \mDelta_2^\T\mW_2 - \mW_1\mX\mX^\T\mDelta_1^\T - \mDelta_1\mX\mX^\T\mW_1^T\|_F^2 = 0
\]
since it follows from \eqref{eq:strict saddle k property} that $\mW_2^\T\mDelta_2 = \mR^\T\widetilde\mW_2^\T \vp_k\valpha^\T\mR = \vzero$ and
\begin{align*}
&\mW_1\mX\mX^\T\mDelta_1^\T = \mR^\T\widetilde\mW_1\mSigma^{-1}\mU^\T\mU\mSigma^2\mU^\T\mU\mSigma^{-1}\vq_k\valpha^\T\mR \\ &= \mR^\T\widetilde\mW_1 \vq_k\valpha^\T\mR= \vzero;
\end{align*}
and
\begin{align*}
& \left\langle \mDelta_2\mDelta_1,(\mW_2 \mW_1 \mX - \mY)\mX^\T \right\rangle \\ &=  \left\langle \vp_k\vq_k^\T\mSigma^{-1}\mU^\T,(\widetilde\mW_2 \widetilde\mW_1 - \mY\mV)\mSigma\mU^\T \right\rangle\\
& = \left\langle \vp_k\vq_k^\T,\widetilde\mW_2 \widetilde\mW_1 \right\rangle - \left\langle \vp_k\vq_k^\T, \mY\mV \right\rangle = - \lambda_k,
\end{align*}
where the last equality utilizes \eqref{eq:strict saddle k property}. Thus, we have
\[
\nabla^2 g(\mZ)[\mDelta,\mDelta] = -2\lambda_k \leq -2\lambda_r.
\]
We finally obtain \eqref{eq:strict saddle 1} by noting that
\begin{align*}
&\|\mDelta\|_F^2 = \|\mDelta_1\|_F^2 + \|\mDelta_2\|_F^2 \\&= \| \vp_k\valpha^\T\mR \|_F^2 + \| \mR^\T\valpha\vq_k^\T \mSigma^{-1}\mU^\T \|_F^2\\
& = 1 + \|\mSigma^{-1} \vq_k\|_F^2 \leq 1 + \|\mSigma^{-1} \|_F^2 \|\vq_k\|_F^2 = 1 + \|\mSigma^{-1} \|_F^2,
\end{align*}
where the inequality follows from the Cauchy-Schwartz inequality $|\va^\T\vb|\leq \|\va\|_2\|\vb\|_2$.

Case $ii$: $r> d_1$. In this case, minimizing $\widetilde g(\widetilde \mZ)$ in \eqref{eq:low rank to YV} is equivalent to finding a low-rank approximation to $\mY\mV$. Let $\Gamma$ denote the indices of the singular vectors $\{\vp_j\}$ and $\{\vq_j\}$ that are included in $\widetilde \mZ$; that is
\[
\left\{\widetilde\vz_i,i\in[d_1]\right\} = \left\{\vzero,\sqrt{\lambda_j}\begin{bmatrix}\vp_j\\\vq_j\end{bmatrix},j\in\Gamma  \right\}.
\]
Then, for any $\widetilde \mZ$, we have
\[
\widetilde\mW_2\widetilde\mW_1 - \mY\mV = \sum_{i\neq \lambda}\lambda_i\vp_i\vq_i
\]
and
\[
\widetilde g(\widetilde \mZ) = \frac{1}{2}\|\widetilde\mW_2\widetilde\mW_1 - \mY\mV\|_F^2 = \frac{1}{2}\left(\sum_{i\neq \Lambda}\lambda_i^2\right),
\]
which implies that $\widetilde \mZ$ is a global minimum of $\widetilde g(\widetilde\mZ)$ if
\[
\|\widetilde\mW_2\widetilde\mW_1 - \mY\mV\|_F^2 = \sum_{i>d_1}\lambda_i^2.
\]
To simply the following analysis, we assume $\lambda_{d_1}> \lambda_{d_1 +1}$; but the argument is similar in the case of repeated eigenvalues at $\lambda_{d_1}$ (i.e., $\lambda_{d_1}= \lambda_{d_1 +1} = \cdots$). In this case, we know for any $\mZ\in \setC \setminus \calX$ that is not a global minimum, there exists $\Omega\subset [r]$ which contains $k\in\Omega,k\leq d_1$ such that
\[
\mY\mV - \widetilde\mW_2\widetilde\mW_1 =\sum_{i\in\Omega}\lambda_i\vp_i\vq_i^\T.
\]

Similar to Case $i$, we have
\e
\vp_k^\T \widetilde \mW_2 = \vzero, \ \widetilde\mW_1 \vq_k = \vzero.
\label{eq:strict saddle k property caseII}\ee
Let $\valpha\in\R^{d_1}$ be the eigenvector associated with the smallest eigenvalue of $\widetilde\mZ^\T\widetilde \mZ$. By the form of $\widetilde\mZ$ in~\eqref{eq:set of critical points}, we have
\begin{align}
\|\widetilde\mW_2 \valpha\|_2^2  = \|\widetilde\mW_1^\T \valpha\|_2^2 \leq \lambda_{d_1+1},
\label{eq:strict saddle alpha caseII}
\end{align}
where the inequality attains equality when $d_1+1\in\Omega$. As in Case $i$, we construct $\mDelta$ by setting $\mDelta_{2} = \vp_k\valpha^\T\mR$ and $\mDelta_{1} = \mR^\T\valpha\vq_k^\T \mSigma^{-1}\mU^\T$. We now show that $\mZ$ is a strict saddle by arguing that $g(\mZ)$ has a strictly negative curvature along the constructed direction $\mDelta$ (i.e., $[\nabla^2g(\mZ)](\mDelta,\mDelta)<0$) by  computing the three terms in \eqref{eq:Hessian} as follows:
\begin{align*}
&\left\|(\mW_2 \mDelta_1 + \mDelta_2\mW_1)\mX_1 \right\|_F^2\\ &= \left\|\widetilde \mW_2\valpha\vq_k^\T \mV^\T + \vp_k\valpha^\T\widetilde\mW_1 \mV^\T   \right\|_F^2\\
& = \left\|\widetilde \mW_2\valpha\right\|_F^2 + \left\| + \valpha^\T\widetilde\mW_1 \right\|_F^2 + 2\left\langle \widetilde \mW_2\valpha\vq_k^\T, \vp_k\valpha^\T\widetilde\mW_1 \right\rangle\\
& \leq 2\lambda_{d_1 +1},
\end{align*}
where the last line follows from \eqref{eq:strict saddle k property caseII} and \eqref{eq:strict saddle alpha caseII};
\begin{align*}
\left\|(\mW_2 \mDelta_1 + \mDelta_2\mW_1)\mX_1 \right\|_F^2 = 0
\end{align*}
holds with a similar argument as in \eqref{eq:third term 0}; and
\begin{align*}
&\left\langle \mDelta_2\mDelta_1,(\mW_2 \mW_1 \mX - \mY)\mX^\T \right\rangle\\ &=  \left\langle \vp_k\vq_k^\T\mSigma^{-1}\mU^\T,(\widetilde\mW_2 \widetilde\mW_1 - \mY\mV)\mSigma\mU^\T \right\rangle\\
& = \left\langle \vp_k\vq_k^\T,\widetilde\mW_2 \widetilde\mW_1 \right\rangle - \left\langle \vp_k\vq_k^\T, \mY\mV \right\rangle \\
&= - \lambda_k \leq -\lambda_{d_1},
\end{align*}
where the last equality used \eqref{eq:strict saddle k property caseII} and the fact that $k\leq d_1$. Thus, we have
\[
\nabla^2 g(\mZ)[\mDelta,\mDelta] \leq -2(\lambda_{d_1} - \lambda_{d_1+1}),
\]
completing the proof of \Cref{lem:local minimum}.


\bibliographystyle{spmpsci}      
\bibliography{nonconvex}

\end{document}